\documentclass[10pt,twocolumn,letterpaper]{article}
\usepackage[pagenumbers]{cvpr}
\usepackage{svg}
\usepackage{multirow}
\usepackage{multicol}

\usepackage{atbegshi}
\usepackage{picture}
\definecolor{cvprblue}{rgb}{0.21,0.49,0.74}
\usepackage[breaklinks,colorlinks,allcolors=cvprblue]{hyperref}
\def\paperID{12}
\def\confName{CVPR}
\def\confYear{2026}
\newcommand{\myheader}{Accepted to IEEE/CVF Conference on Computer Vision
and Pattern Recognition (CVPR) Workshops, 2026}
\newcommand{\mycopyrightnotice}{%
  \begin{minipage}{14cm}
    \scriptsize © 2026 IEEE. Personal use of this material is permitted.
    Permission from IEEE must be obtained for all other uses, in any current
    or future media, including reprinting/republishing this material for
    advertising or promotional purposes, creating new collective works, for
    resale or redistribution to servers or lists, or reuse of any copyrighted
    component of this work in other works.
  \end{minipage}%
}
\title{DinoRADE: Full Spectral Radar-Camera Fusion with Vision Foundation
Model Features for Multi-class Object Detection in Adverse Weather}
\author{
Christof Leitgeb$^{1,2}$ \quad
Thomas Puchleitner$^{1}$ \quad
Max Peter Ronecker$^{2,3}$ \quad
Daniel Watzenig$^{2,3}$ \\
$^{1}$Infineon Technologies AG, Austria \quad
$^{2}$Graz University of Technology, Austria \\
$^{3}$Virtual Vehicle Research GmbH, Austria \\
{\tt\small \{christof.leitgeb, thomas.puchleitner\}@infineon.com} \\
{\tt\small max.ronecker@v2c2.at \quad daniel.watzenig@tugraz.at}
}
\begin{document}

\newif\iffirstpage
\firstpagetrue

\AtBeginShipout{%
  \AtBeginShipoutUpperLeft{%
    \unitlength=1pt
    \begin{picture}(0,0)
      \put(52,-20){\parbox{\textwidth}{\scriptsize\myheader}}
    \end{picture}
  }%
  \iffirstpage
    \AtBeginShipoutUpperLeft{%
      \unitlength=1pt
      \begin{picture}(0,0)
        \put(52,-774){\mycopyrightnotice}
      \end{picture}
    }%
    \global\firstpagefalse
  \fi
}

\maketitle
\begin{abstract}
Reliable and weather-robust perception systems are essential for safe autonomous driving and typically employ multi-modal sensor configurations to achieve comprehensive environmental awareness. While recent automotive FMCW Radar-based approaches achieved remarkable performance on detection tasks in adverse weather conditions, they exhibited limitations in resolving fine-grained spatial details particularly critical for detecting smaller and vulnerable road users (VRUs). Furthermore, existing research has not adequately addressed VRU detection in adverse weather datasets such as K-Radar. We present DinoRADE, a Radar-centered detection pipeline that processes dense Radar tensors and aggregates vision features around transformed reference points in the camera perspective via deformable cross-attention. Vision features are provided by a DINOv3 Vision Foundation Model. We present a comprehensive performance evaluation on the K-Radar dataset in all weather conditions and are among the first to report detection performance individually for five object classes. Additionally, we compare our method with existing single-class detection approaches and outperform recent Radar-camera approaches by 12.1\%. The code is available under https://github.com/chr-is-tof/RADE-Net.
\end{abstract}
\section{Introduction}
\label{sec:intro}
Safe autonomous driving requires robust and reliable perception systems to interact with other road users and navigate complex scenarios, often in adverse weather conditions like fog, rain, and snow, which additionally disturb the view of optical sensors~\cite{feng_review_weather_2025}. To tackle these challenges, automotive systems must rely on a variety of sensor technologies to guarantee reliable perception. Cameras provide high structural details which contribute to reliable object classification, but they lack inherent depth information which can lead to depth estimation errors, especially for texture-less regions, reflective surfaces, varying illumination conditions or perspective illusions~\cite{simeonov_realtime_2026}. LiDAR sensors combine high structural information with accurate depth information but rely on expensive hardware and are limited in adverse weather conditions, where atmospheric particles like smoke, fog, rain, and snow block their relatively short wavelength signal~\cite{kurup_dsor_2021, feng_review_weather_2025}. In contrast, Radar sensors emit a signal with significantly longer wavelength, which results in a constant perception performance across various adverse weather conditions~\cite{fent_dpft_2025, leitgeb_rade-net_2025}. However, while providing relatively accurate distance information, the large antenna array structures result in low angular resolution, which makes precise object separation challenging, especially for small objects like pedestrians and cyclists~\cite{yao_exploring_2025}. Another significant benefit of Radar sensors is the direct measurement of the relative radial velocity via the Doppler effect. This information can be employed for tasks like object separation and motion classification~\cite{yao_exploring_2025}.  Additionally, the full Doppler spectrogram contains characteristic micro-Doppler signatures caused by micro-motions, such as rotation and vibration from object parts, which are unique for different types of road users and motions~\cite{yao_exploring_2025}. These signatures contain rich information for human and vehicle motion classification, which subsequently benefits close object separation~\cite{yao_exploring_2025, senigagliesi_case_study_2022}.

Given the complementary strengths and weaknesses of different automotive sensors, it becomes evident that a combination of modalities will benefit robust perception. The fusion of Radar and camera combines fine structural details with reliable performance in adverse weather conditions, direct position, and velocity measurements as well as low costs~\cite{feng_review_weather_2025}. Consequently, Radar-camera approaches have been widely explored for object detection and segmentation tasks and reported results demonstrate superior performance~\cite{fent_dpft_2025, schramm_bevcar_2024, xiao_radarcam_2025, li_bevformer_2025, lin_rcbevdet_2024}. However, while visual images from a camera are fairly compact, raw Radar data consists of sampled time-series data which contains frequency and phase information about objects in the scene and comes with large data sizes for each single frame~\cite{yao_exploring_2025}. To obtain a portable format, they are usually Fast-Fourier transformed (FFT) and reduced by adaptive thresholding techniques, which results in a 3-dimensional Radar point cloud with additional Doppler values. However, this process results in a significant loss of structural detail, yielding only few data points per object, especially for smaller classes like pedestrians and cyclists. To counteract data sparsity, recent approaches incorporate denser Radar information extracted from lower-level processing stages. Examples include raw time-series analog-to-digital (ADC) converted data~\cite{yang_adcnet_2023, giroux_t-fftradnet_2023}, sparse FFT-processed range-azimuth-Doppler-elevation (RADE) tensors~\cite{kong_rtnh+_2025, paek_kradar_2022} as well as 2D and 3D projections from RADE tensors~\cite{rebut_fft-radnet_2022,fent_dpft_2025, leitgeb_rade-net_2025,guan_waveletbased_2026}.

Conventional vision-based perception pipelines utilize task-specific architectures, such as ResNet~\cite{he_resnet_2016} and YOLO~\cite{redmon_yolo_2016}, which are often initialized with pre-trained weights or trained from scratch to accommodate domain-specific requirements. However, the advancement of Vision Foundation Models (VFMs)~\cite{oquab_dinov2_2023, simeoni_dinov3_2025} has introduced a paradigm shift by providing general-purpose pre-trained feature extractors that demonstrate superior generalization capabilities. These models can extract semantically meaningful representations from visual scenes, including rare scenarios potentially absent from task-specific training datasets. Consequently, VFMs have demonstrated significant efficacy in automotive perception systems, offering enhanced robustness and adaptability compared to conventional approaches.

Many state-of-the-art Radar-camera or Radar-only approaches are evaluated on datasets without severe weather conditions like fog or snow~\cite{yao_com_review_2024,caesar_nuscenes_2020}, thereby limiting the full potential of Radar sensing. Conversely, approaches evaluated on adverse weather datasets like K-Radar~\cite{paek_kradar_2022} predominantly focus on vehicle detection~\cite{fent_dpft_2025, paek_kradar_2022, kong_rtnh+_2025, liu_echoes_beyond_2023}, thus missing the effect of high resolution visual details that might be crucial for VRU detection and classification. To the best of our knowledge, we are the first to combine dense Radar representations derived from 3D FFT tensor projections with high-resolution semantic features extracted from Vision Foundation Models, while providing comprehensive evaluations across diverse weather conditions and multiple object classes including VRUs. This enables full exploitation of the complementary strengths of both sensing modalities. Specifically, our contributions are as follows:
\begin{itemize}
    \item A weighted query lifting method to refine Radar queries based on their true distribution in the range-azimuth-elevation projection and aggregate inter-perspective features from synchronized DINOv3~\cite{simeoni_dinov3_2025} VFM representations.
    \item Adaptive fusion strategy to selectively refine features from RADE-Net~\cite{leitgeb_rade-net_2025} Radar backbone to increase detection performance with special focus on smaller object classes.
    \item Comprehensive performance evaluation in adverse weather conditions and individually for 5 object classes of the K-Radar~\cite{paek_kradar_2022} dataset.
\end{itemize}

\section{Background and Related Works}
\label{sec:background}
\subsection{Automotive Camera-Radar Datasets}
Automotive datasets, which provide camera images and Radar data, have strong variations in size, class categories, use of sensor modalities, data representation, environment scenarios, and annotation quality~\cite{yao_com_review_2024}. While camera data is almost exclusively represented in RGB format, with the exception of RaDICal~\cite{lim_radical_2021} providing RGB-depth data, Radar experiences strong variations in data representation. RaDICal and RADIal~\cite{rebut_radial_2022} provide raw time-series data. CARRADA~\cite{ouaknine_carrada_2021}, UWCR~\cite{xiangyu_uwcr_2022}, and RADDet~\cite{zhang_raddet_2021} offer processed range-azimuth-Doppler (RAD) tensors.  K-Radar~\cite{paek_kradar_2022} provides higher resolution range-azimuth-Doppler-elevation (RADE) tensors. Furthermore, very prominent automotive datasets like nuScenes~\cite{caesar_nuscenes_2020}, View-of-Delft~\cite{palffy_vod_dataset_2022}, and Astyx~\cite{meyer_astyx_2019} only provide sparse Radar point clouds, but instead include up to 23 different object classes along with high quality 3D bounding box annotations. Most time-series and FFT tensor-based datasets lack annotation quality with the exception of K-Radar, which provides 3D rotated bounding boxes for 7 object classes. To our knowledge, K-Radar~\cite{paek_kradar_2022} is the only large-scale Radar-camera dataset capturing 7 weather conditions, 8 road types, and day/night variations, making it suitable for exploring the complementary advantages of Radar and camera across diverse scenarios.

\subsection{Radar-only Detection}
Radar-only object detection and segmentation approaches can be grouped in three primary categories based on their underlying data representation~\cite{yao_exploring_2025}: Raw time-series ADC data~\cite{yang_adcnet_2023, giroux_t-fftradnet_2023}, FFT-processed spectral data~\cite{rebut_fft-radnet_2022, leitgeb_rade-net_2025, paek_kradar_2022, kong_rtnh+_2025}, and sparse point cloud data~\cite{musiat_radarpillars_2024, palffy_vod_dataset_2022}. ADCNet~\cite{yang_adcnet_2023} uses a distillation method to learn RAD tensors from raw ADC data and optimize the network to predict the range-azimuth coordinate of objects in the scene. Similarly, T-FFTRadNet~\cite{giroux_t-fftradnet_2023} learns the transformation from ADC to FFT representation and utilizes hierarchical Swin Vision transformers on the patched FFT data. FFT-RadNet~\cite{rebut_fft-radnet_2022} reduces memory requirements by recovering angular information from the range-Doppler (RD) spectrum. RADE-Net~\cite{leitgeb_rade-net_2025} creates 3D projections from RADE tensors which preserve rich Doppler and elevation features, thus achieving superior performance. RTNH~\cite{paek_kradar_2022} and RTNH+~\cite{kong_rtnh+_2025} employ a sparse RADE tensor by reducing the full tensor to the top 10\% of power measurements. RadarPillars~\cite{musiat_radarpillars_2024} uses a pillar-based approach similar to LiDAR processing and introduces self-attention as well as radial velocity encoding to handle sparsity of Radar data, thus surpassing the Radar baseline on the VoD dataset~\cite{palffy_vod_dataset_2022}.

\subsection{Camera-only Detection}
3D object detection from monocular RGB-cameras presents a challenge due to the lack of inherent depth information, which requires network architectures to recover depth information from perspective features in the image~\cite{jin_survey_2024}. Image-based 3D detectors can be categorized based on 2D or 3D features and further grouped into result lifting, feature lifting, or data lifting methods~\cite{ma_survey-images_2024}. First, result lifting methods use 2D features to estimate the location of objects in the image plane and then lift them into 3D. Therefore, they are similar to classical 2D detectors, which include region-based methods like R-CNN~\cite{girshick_rcnn_2014} and single-shot methods like CenterNet~\cite{zhou_objects-as-points_2019}. Second, feature lifting methods generate 3D features from lifted 2D features and predict the result in 3D. Examples like BEVDet~\cite{huang_bevdet_2022} collapse the 3D features to BEV before predicting the final result. Last, data lifting methods lift the input data from 2D to 3D and generate the results directly. This concept is widely associated with pseudo-LiDAR approaches~\cite{wang_pseudo-lidar_2019} where image-based depth maps are converted to pseudo-LiDAR representations, which allows to apply existing LiDAR-based detection algorithms~\cite{ma_survey-images_2024, wang_pseudo-lidar_2019}.

\subsection{Vision Foundation Models}
Recent advancements in foundation models have demonstrated remarkable performance in a variety of computer vision and autonomous driving applications, including object detection, classification, and segmentation~\cite{oquab_dinov2_2023, simeoni_dinov3_2025, chen_vit_adapter_2022, lu_vfm_survey_2025}. VFMs are large scale models pre-trained on massive, diverse datasets which enables them to capture complex patterns and features from unseen images and provide a reusable representation across various tasks and domains~\cite{lu_vfm_survey_2025}.  

\subsection{Radar-Camera Fusion}
The fusion of Radar and camera combines the most complementary features of both modalities which makes it a popular research direction for perception in adverse weather~\cite{feng_review_weather_2025}. DPFT~\cite{fent_dpft_2025} creates multi-view projections from Radar FFT data in range-azimuth and azimuth-elevation view. Both Radar and camera are processed by ResNet~\cite{he_resnet_2016} backbones, view-transformed and fused using deformable attention~\cite{zhu_deformable-detr_2020}. BEVCar~\cite{schramm_bevcar_2024} encodes surround-view images using a DINOv2~\cite{oquab_dinov2_2023} pretrained VFM in combination with a vision transformer adapter~\cite{chen_vit_adapter_2022}. Radar is encoded from point-cloud format followed by deformable self- and cross-attention~\cite{zhu_deformable-detr_2020} where the cross-attention takes lifted 3D queries from the VFM output. WRCFormer~\cite{guan_waveletbased_2026} utilizes multi-view Radar encoders in range-azimuth and elevation-azimuth perspectives which are subsequently decomposed by a Wavelet-transform based attention module and adaptively fused.

\begin{figure*}[ht]
  \centering
  \includegraphics[width=0.8\linewidth]{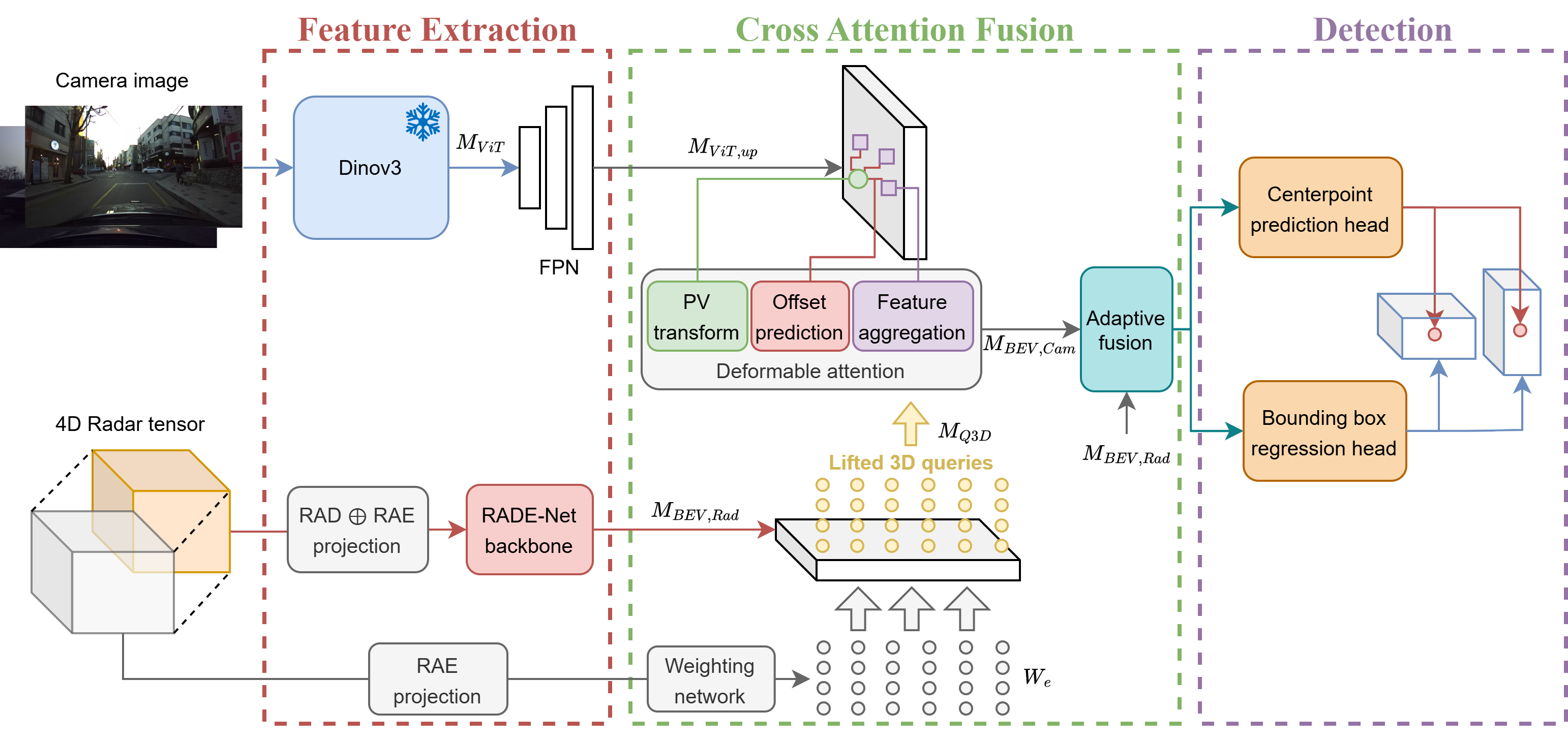}
  \caption{Overview of the DinoRADE architecture.}
  \label{fig:architecture_overview}
\end{figure*}

\begin{figure}[ht]
  \centering
  \includegraphics[width=\linewidth]{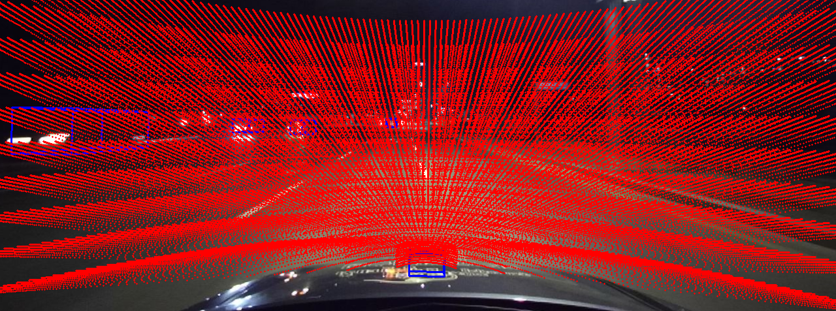}
  \caption{Reference points projected from 3D Radar queries to camera image (red) and Radar sensor location (blue).}
  \label{fig:reference_points}
\end{figure}

\section{Methodology}
\label{sec:methodology}
Our architecture in Fig.~\ref{fig:architecture_overview} is designed to extract 3D positional and spectral Doppler features from Radar 3D projections and aggregate structural details from inter-perspective VFM features. The RAD and RAE projections are processed utilizing a modified RADE-Net backbone from~\cite{leitgeb_rade-net_2025}, yielding a BEV feature map that preserves the inherent range-azimuth coordinates of Radar data while Doppler and elevation information is encoded into the 128 feature channels. Subsequently, the BEV features are lifted to receive a 3D query representation, which is further refined through elevation weights, learned from the spectral distribution of the RAE projection. The query positions provide 3D information while removing the need for computationally expensive 3D CNNs during feature extraction~\cite{song_3d-2d-cnn_2020}.

The camera images are processed utilizing a pretrained DINOv3 ViT-S/16~\cite{simeoni_dinov3_2025}, which extracts generalized image features that are subsequently up-sampled, thus resulting in a perspective-view (PV) feature map. To aggregate features in the PV map with query information from the BEV map, we utilize deformable attention~\cite{zhu_deformable-detr_2020}. With known transformation matrices, each 3D query position can be mapped to a 2D location in the PV feature map. To exploit this relationship, we transform a set of reference points, visualized in Fig.~\ref{fig:reference_points}, for all queries from Radar to camera PV coordinates. Furthermore, an offset prediction network predicts offsets for each 3D query to learn where to aggregate features in the PV feature map as described in~\cite{zhu_deformable-detr_2020}. The aggregated features are utilized to update the BEV Feature map with refined visual information. Camera images are severely affected by adverse weather conditions, resulting in occlusions, noise and obstructed views. This leads to the PV feature map lacking informative content for the BEV queries to aggregate from. To address this limitation, we introduce a gated residual fusion method, which enables the model to selectively refine the BEV feature map with visual information or rely exclusively on Radar data, depending on the quality, robustness and relevance of the visual input.

After adaptive fusion, the 3D features are again mapped onto a BEV map, which is then forwarded to a detection head adapted from~\cite{leitgeb_rade-net_2025}.
Specifically, we modified the focal loss to better match the true shape and size of different object bounding boxes in the range-azimuth format of the feature maps. This results in better detection performance especially for smaller road users like pedestrians and cyclists. 

\subsection{Radar Feature Extraction}
We adapt the RADE-Net backbone~\cite{leitgeb_rade-net_2025} to employ a dual encoding architecture in which both 3D projections undergo independent processing before concatenation along the feature dimension. This approach allows the model to selectively integrate Doppler and elevation information prior to feature extraction, thereby enhancing flexibility when supplementary elevation data is introduced during the subsequent lifting stage. Upon completion of processing, the backbone generates a BEV feature map $M_{BEV,Rad}\in \mathbb{R}^{256\times112\times128}$, which encodes each bin in the 256×112 Radar range-azimuth domain with a 128-dimensional feature representation.

\subsection{Camera Feature Extraction}
We employ a DINOv3 ViT-S/16~\cite{simeoni_dinov3_2025} pretrained on 1.689 billion images to extract general-purpose visual feature representations. The input image, with a resolution of 720x1280 pixels, is divided into non-overlapping 16x16 pixel patches, yielding 3600 patches in total. These patches are processed by the frozen vision transformer, which produces an output feature map $M_{ViT}\in \mathbb{R}^{45\times80\times 384}$. To obtain features at a higher spatial resolution, the output feature map undergoes spatial upsampling using a two-layer Feature Pyramid Network (FPN) which reduces the feature dimension, resulting in an upsampled feature map $M_{ViT,up}\in \mathbb{R}^{180\times320\times 128}$.

\subsection{Weighted Feature Lifting}
To account for the distribution of Radar features along the elevation dimension, we expand the BEV feature map into E=10 elevation segments and apply appropriate weighting to the bins within each segment. This yields the lifted 3D query map $M_{Q3D}\in \mathbb{R}^{256\times112\times128\times10}$ according to:
\begin{equation}    
    M_{Q3D}=[M_{BEV,Rad}]_{i=1}^{E}\times W_e^T
\end{equation}
where the weights $W_e$ are learned from the RAE projection.
\begin{equation}
    W_e = [\operatorname{Softmax}(\operatorname{MLP}(\mathcal{P}^{RAE}))]_{i=1}^{128}
\end{equation}
This concept exploits the property that the RAE projection $\mathcal{P}^{RAE}$ provides a direct representation of the spectral power distribution along the elevation dimension for each range-azimuth bin within the feature map.

\subsection{Deformable Cross Attention}
The cross attention module employs deformable attention~\cite{zhu_deformable-detr_2020} to aggregate camera features for a set of 3D queries derived from Radar input. To guide this aggregation process, we generate reference points for each 3D query and project them onto $M_{ViT,up}$. This projection leverages Radar sampling parameters, sensor positions, and camera intrinsics/extrinsics from~\cite{paek_kradar_2022}, providing spatial priors that indicate relevant regions in the perspective view. Figure~\ref{fig:reference_points} illustrates the resulting reference point distribution on the camera image. Following~\cite{zhu_deformable-detr_2020}, we predict four offsets around each reference point to sample features, which are then aggregated to update $M_{Q3D}$. Finally, we average the updated 3D query map along the height dimension to obtain the BEV refined map $M_{BEV,Cam}$.

\subsection{Adaptive Fusion}
The adaptive fusion model is designed to learn, for each range-azimuth bin, whether features should be extracted from the Radar-only feature map $M_{BEV,Rad}$ or from the VFM-refined feature map $M_{BEV,Cam}$. Specifically, we employ a gated fusion approach in which a learned gate $\Gamma \in \mathbb{R}^{256\times112\times128}$ dynamically regulates the composition of the fused Radar-camera feature map $M_{f}$ as follows:
\begin{equation}
    M_{f} = \Gamma M_{BEV,Rad} + (1-\Gamma) M_{BEV,Cam}
\end{equation}
with
\begin{equation}
    \Gamma = \operatorname{Sigmoid(}\operatorname{MLP}(M_{BEV,Rad}\oplus M_{BEV,Cam}))
\end{equation}
where $\oplus$ represents the concatenation operation and MLP a multilayer perceptron with two layers.

\subsection{Detection Head}
The detection head performs center-point detection, object classification, and bounding box regression based on the fused BEV feature map $M_f$, thereby exploiting the Radar's native range-azimuth coordinate representation~\cite{leitgeb_rade-net_2025}. Multi-class detection is achieved through class-specific heatmaps, with each object class represented in a separate channel.

\subsection{Loss}
We adapt the loss strategy from~\cite{leitgeb_rade-net_2025} by employing a combination of loss components: a focal loss for center-point convergence, and a composite regression loss that integrates the Gaussian-Wasserstein Distance (GWD)~\cite{xue_gwd-loss_2021} with a smooth L1 term. While the focal loss performs adequately for larger objects such as cars, busses and trucks, it presents challenges for smaller objects like pedestrians and cyclists. The isotropic Gaussian distribution generated at the ground truth center point in range-azimuth coordinates with a fixed $\sigma=3$ covers a disproportionately large physical area at greater range values due to the polar coordinates of the Radar data. This issue is particularly problematic for small objects, as the model struggles to converge to the true object center. We conducted an analysis of various $\sigma$ values and determined that $\sigma=0.75$ yields optimal center-point convergence across all object classes while simultaneously enhancing the detection performance for smaller road users.
\begin{figure*}[htbp]
    \centering
    \includegraphics[width=\linewidth]{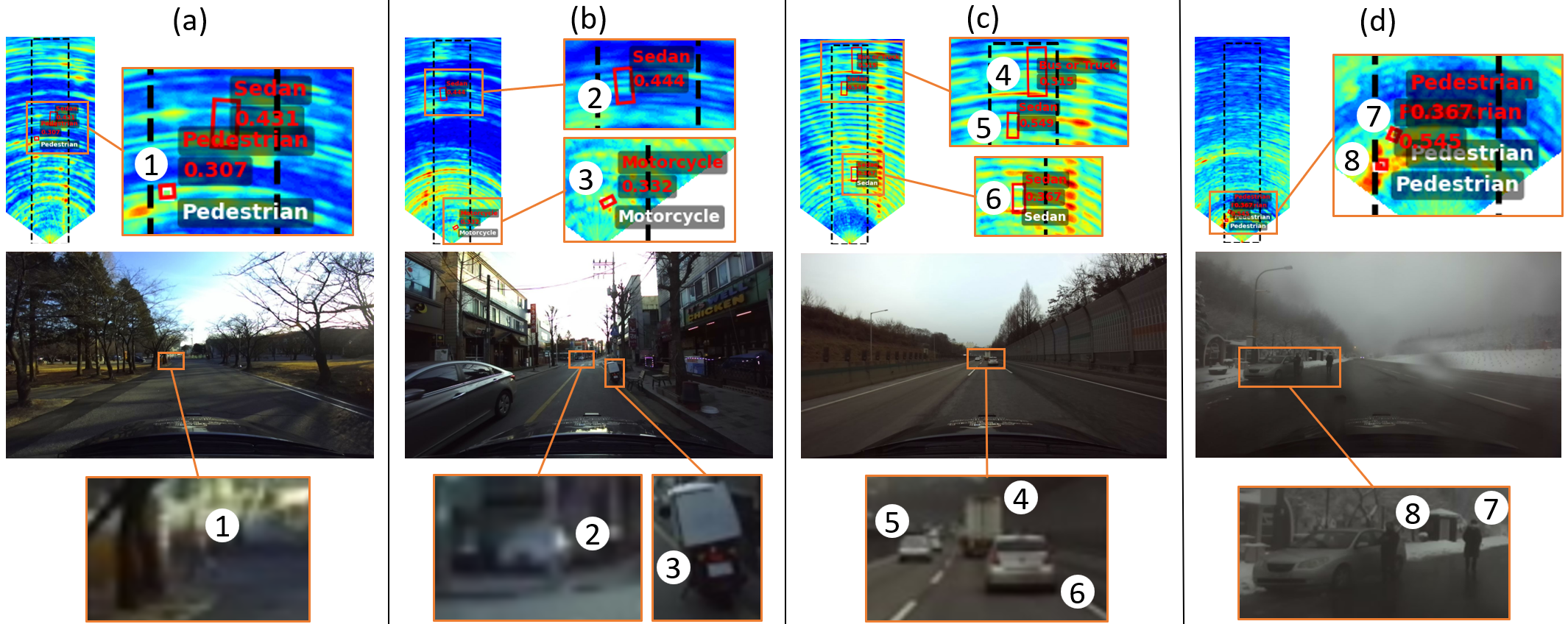}
    \caption{DinoRADE performance visualization in four different scenarios: (a) university campus, (b) alleyway, (c) highway, and (d) road shoulder. Ground truth bounding boxes and class labels are shown in white. Predicted bounding boxes with predicted class and confidence score are shown in red. The ROI where detections are considered is marked with a dashed black rectangle. Magnified views are indicated in orange.}
    \label{fig:model_vis}
\end{figure*}

\begin{table*}[htbp]
    \centering
    \caption{3D Average Precision (AP) comparison on 'Sedan' class in all weather conditions using K-Radar v1.1.}
    \begin{tabular}{l c cccccccc}
    \toprule
        Methods & Modality & Total & Normal & Overcast & Fog & Rain & Sleet & Light Snow & Heavy Snow\\
        \midrule
        Voxel-RCNN~\cite{deng_voxelrcnn_2021} & \multirow{3}{*}{L} & 46.4 & 81.8 & 69.6 & 48.8 & 47.1 & 46.9 & 54.8 & 37.2\\
        CasA~\cite{wu_casa_2022} & & 50.9 & \underline{82.2} & 65.6 & 44.4 & 53.7 & 49.9 & 62.7 & 36.9 \\
        TED-S~\cite{wu_teds_2023} &  & 51.0 & 74.3 & 68.8 & 45.7 & 53.6 & 44.8 & 63.4 & 36.7 \\
        \midrule
        RTNH~\cite{paek_kradar_2022} & \multirow{3}{*}{R} & 47.4 & 49.9 & 56.7 & 52.8 & 42.0 & 41.5 & 50.6 & 44.5\\
        RTNH+~\cite{kong_rtnh+_2025} & & 57.6 & - & - & - & - & - & - & -\\
        RADE-Net~\cite{leitgeb_rade-net_2025} & & \underline{66.7} & 66.3 & 74.5 & \textbf{82.1} & \underline{57.2} & \textbf{69.4} & 63.9 & \textbf{68.9}\\ 
        \midrule
        VPFNet~\cite{zhu_vpfnet_2023} & \multirow{3}{*}{LC} & 52.2 & 81.2 & 76.3 & 46.3 & 53.7 & 44.9 & 63.1 & 36.9 \\
        TED-M~\cite{wu_teds_2023} & & 52.3 & 77.2 & 69.7 & 47.4 & 54.3 & 45.2 & 64.3 & 36.3 \\
        MixedFusion~\cite{zhang_mixedfusion_2025} & & 55.1 & \textbf{84.5} & \underline{76.6} & 53.3 & 55.3 & 49.6 & \underline{68.7} & 44.9 \\
        \midrule
        EchoFusion~\cite{liu_echoes_beyond_2023} & \multirow{4}{*}{RC} & 47.4 & 51.5 & 65.4 & 55.0 & 43.2 & 14.2 & 53.4 & 40.2 \\
        DPFT~\cite{fent_dpft_2025} & & 56.1 & 55.7 & 59.4 & 63.1 & 49.0 & 51.6 & 50.5 & 50.5 \\
        WRCFormer~\cite{guan_waveletbased_2026} & & 58.7 & 55.9 & 59.7 & \underline{70.1} & 52.2 & 54.7 & 59.2 & 54.0 \\
        DinoRADE (ours) & & \textbf{70.8} & 72.8 & \textbf{84.9} & 69.7 & \textbf{68.4} & \underline{58.4} & \textbf{70.4} & \underline{65.0}\\
        \bottomrule
    \end{tabular}
    \label{tab:pc_sedan}
\end{table*}

\begin{table*}[htbp]
    \centering
    \caption{AP and mAP (Total) evaluation of all road users in all weather conditions using K-Radar v2.1. (-) indicates no representation in the test set and (*) indicates a severe under-representation of under 2\% of total objects in the train set of the respective weather condition.}
    \begin{tabular}{l cc cc cc cc cc cc}
    \toprule
        \multirow{2.5}{*}{\begin{tabular}{@{}c@{}} Weather \end{tabular}} & \multicolumn{2}{c}{Total} & \multicolumn{2}{c}{Sedan} & \multicolumn{2}{c}{Bus or Truck} & \multicolumn{2}{c}{Pedestrian} & \multicolumn{2}{c}{Motorcycle} & \multicolumn{2}{c}{Bicycle}\\
        \cmidrule(lr){2-3} \cmidrule(lr){4-5} \cmidrule(lr){6-7} \cmidrule(lr){8-9} \cmidrule(lr){10-11} \cmidrule(lr){12-13}
        & 3D & BEV & 3D & BEV & 3D & BEV & 3D & BEV & 3D & BEV & 3D & BEV \\
        \midrule
        Total  & 36.99 & 39.61 & 71.38 & 75.32 & 54.92 & 58.93 & 33.01 & 38.12 & \phantom{*}3.77 & \phantom{*}3.77 & 21.89 & 21.89 \\
        Normal & 29.04 & 30.08 & 71.52 & 74.13 & 46.61 & 49.17 & *0.99 & *0.99 & *3.77 & *3.77 & 22.32 & 22.32 \\
Overcast & 70.89 & 74.33 & 74.20 & 75.16 & 67.57 & 73.51 & - & - & - & - & - & - \\
Fog & 67.41 & 75.68 & 83.84 & 89.91 & - & - & 50.98 & 61.46 & - & - & - & - \\
Rain & 24.10 & 25.97 & 71.32 & 76.92 & *0.99 & *0.99 & *0.00 & *0.00 & - & - & - & - \\
Sleet & 47.14 & 49.63 & 66.89 & 70.43 & 44.69 & 46.41 & 29.85 & 32.04 & - & - & - & - \\
Light Snow & 70.73 & 76.63 & 69.29 & 76.72 & 72.17 & 76.55 & - & - & - & - & - & - \\
Heavy Snow & 71.93 & 76.34 & 66.43 & 66.75 & 77.42 & 85.94 & - & - & - & - & - & - \\
 \bottomrule
    \end{tabular}
    \label{tab:performance_multiobject_multiweather}
\end{table*}

\begin{table*}[htbp]
    \centering
    \setlength{\tabcolsep}{4pt}
    \caption{AP and mAP (Total) comparison with Radar-Camera and Radar-only methods for multiple object classes using K-Radar v2.1.}
    \begin{tabular}{l c cc cc cc cc cc cc}
    \toprule
        \multirow{2.5}{*}{\begin{tabular}{@{}c@{}} Methods \end{tabular}} & \multirow{2.5}{*}{Mod.} & \multicolumn{2}{c}{Total} & \multicolumn{2}{c}{Sedan} & \multicolumn{2}{c}{Bus or Truck} & \multicolumn{2}{c}{Pedestrian} & \multicolumn{2}{c}{Motorcycle} & \multicolumn{2}{c}{Bicycle}\\
        \cmidrule(lr){3-4} \cmidrule(lr){5-6} \cmidrule(lr){7-8} \cmidrule(lr){9-10} \cmidrule(lr){11-12} \cmidrule(lr){13-14}
        & & 3D & BEV & 3D & BEV & 3D & BEV & 3D & BEV & 3D & BEV & 3D & BEV \\
        \midrule
        RADE-Net~\cite{leitgeb_rade-net_2025} & R  & 20.93 & 23.96 & 56.75 & 63.21 & 40.99 & 48.29 & 5.40  & 6.79  & 0.00  & 0.00  & 1.49     & 1.49     \\
        DPFT~\cite{fent_dpft_2025} & RC & 11.65 & 13.54 & 38.99 & 41.02 & 17.60 & 20.55 & 1.66 & 6.14  & 0.00  & 0.00  & 0.00 & 0.00     \\
        Ours & RC & 36.99 & 39.61 & 71.38 & 75.32 & 54.92 & 58.93 & 33.01 & 38.12 & 3.77 & 3.77 & 21.89 & 21.89 \\
    \bottomrule
    \end{tabular}
    \label{tab:performance_multiclass_noweather}
\end{table*}

\section{Experiments}
\label{sec:experiments}
We conducted a comprehensive experimental evaluation of our approach, with particular emphasis on the detection of vulnerable road users and demonstrate the robustness of our method by evaluating Average Precision (AP) and mean Average Precision (mAP) metrics across various weather conditions. Following the K-Radar benchmark~\cite{kong_rtnh+_2025}, we evaluate our trained model on single-class detection for the 'Sedan' class, which exhibits strong representation across all weather conditions in the dataset and thus serves as a reliable performance indicator. We employ this single-class evaluation to compare our approach against other methods utilizing different sensor modality configurations, as presented in Table~\ref{tab:pc_sedan}.

To provide more comprehensive insights, we present the AP and mAP detection performance for different road user classes across all weather conditions in Table~\ref{tab:performance_multiobject_multiweather}. Following the K-Radar benchmark~\cite{paek_kradar_2022}, existing works primarily report AP performance for the 'Sedan' and 'Bus or Truck' classes which are represented strongly in the dataset~\cite{paek_kradar_2022, song_enhanced-diverse_2026}. To our knowledge, only~\cite{leitgeb_rade-net_2025} additionally evaluates detection performance on 'Pedestrian' and 'Bicycle' classes, limiting direct comparison for multiclass detection. To facilitate a more comprehensive evaluation, we establish an additional baseline by retraining DPFT~\cite{fent_dpft_2025} and extending the single-class detection approach to all five object classes, with results presented in Table~\ref{tab:performance_multiclass_noweather}. Following K-Radar~\cite{paek_kradar_2022}, we consider detections within a Region of Interest (ROI) of $x\in[0,72m]$, $y\in [-6.4m, 6.4m]$ and $z\in [-2m, 6m]$.

\subsection{Dataset}
Our evaluation is conducted on the large-scale K-Radar dataset~\cite{paek_kradar_2022} and adheres to the official KITTI evaluation protocol~\cite{geiger_kitti_2012}. Results reported in Table~\ref{tab:pc_sedan} are derived from the earlier dataset label version 1.1~\cite{paek_kradar_2022} to ensure comparability with existing approaches. Table~\ref{tab:performance_multiobject_multiweather}--~\ref{tab:ablation_study} employ the updated labels from v2.1, which include additional annotations missing in v1.1~\cite{sun_k-radar-labeling_2024}. The dataset provides 35k Radar frames along with synchronized camera images. Annotations consist of rotated 3D bounding boxes with seven class labels, of which we utilize five: 'Sedan', 'Bus or Truck', 'Pedestrian', 'Bicycle', and 'Motorcycle'. The additional classes 'Pedestrian Group' and 'Bicycle Group' are severely underrepresented in the dataset and introduce classification ambiguity between the relatively similar 'Pedestrian' and 'Bicycle' classes.
\subsubsection{Training}
We train our model on a NVIDIA A30 GPU for 11 epochs using a batch size of 6. For optimization, we employ the AdamW optimizer with an initial learning rate of 0.001 and a weight decay of 0.01. We apply a cosine annealing learning rate scheduler with a minimum learning rate of $10^{-4}$. Our model has a total number of 31M learnable parameters and 21M frozen weights of the DINOv3 ViT-S/16~\cite{simeoni_dinov3_2025}.
\subsubsection{Inference}
Inference speed was evaluated on an NVIDIA A30 GPU, achieving an average processing time of $190.34$ ms per frame. It is worth noting that the framework has not been optimized for inference speed, indicating substantial potential for improvement and further acceleration in real-time deployment scenarios.
\subsection{Results}
We present qualitative results in Fig.~\ref{fig:model_vis} across four scenarios for different road users. Scenario (a) demonstrates the correct detection of a pedestrian (1) at 37m from the ego vehicle, right next to a false positive (FP) prediction. In scenario (b), while a parked motorcycle (3) is accurately detected, an unlabeled parked vehicle (2) causes a FP prediction. Scenario (c) exhibits similar behavior, where a vehicle (6) traveling ahead is correctly classified, whereas the unlabeled car (5) and truck (4) preceding it result in FP predictions due to missing annotations in the dataset. Finally, scenario (d) demonstrates the successful detection of two pedestrians (7,8), with the adjacent vehicle positioned beyond the ROI.

Examining quantitative results in Table~\ref{tab:pc_sedan}, we observe that purely optical methods utilizing LiDAR and camera demonstrate strong performance under normal and overcast weather conditions but exhibit substantial degradation in adverse weather. In contrast, Radar-only performance remains relatively consistent across all weather conditions, while Radar-camera fusion enhances performance under normal and overcast conditions while maintaining robustness in adverse weather. Notably, our approach outperforms all prior methods in total AP as well as in overcast, rain, and light snow conditions. Additionally, it can be observed that Radar-based methods perform slightly better in adverse weather compared to normal conditions, which may appear counterintuitive but is attributed to the distribution of labeled bounding boxes across weather conditions and detection distances, as discussed in~\cite{huang_l4dr_2025}.

Results presented in Table~\ref{tab:performance_multiobject_multiweather} show the detection performance for different road users across weather conditions. Pedestrian detection performs strong in fog and sleet, attributed to simple scenes, well-represented in both train and test set. Road scenarios in normal conditions are more challenging, including a high number of different road users and complex moving paths. This introduces classification mismatches, for instance where the model mistakes a person standing on a scooter for a pedestrian which generates a false positive in the 'Pedestrian' class and a false negative in the 'Motorcycle' class. Additionally, pedestrians are strongly underrepresented in the training set with 1.6\% in 'normal' and 0\% in 'rain' conditions while representing 28.4\% in sleet, 24\% in fog and 6.9\% in light snow. Similarly, 'Motorcycle' and 'Bicycle' class only represent 1\% and 3.1\% respectively of training data in normal weather condition.

In Table~\ref{tab:performance_multiclass_noweather} we compare our model with RADE-Net~\cite{leitgeb_rade-net_2025} and DPFT~\cite{fent_dpft_2025}. Our method achieves substantially higher performance across multiple classes. However, it should be noted that DPFT was originally designed and optimized for single-class detection rather than multi-class scenarios. Consequently, the multi-class evaluation may not fully reflect DPFT's capabilities within its intended operational scope.

\subsection{Ablation Study}
In Table~\ref{tab:ablation_study} we demonstrate the performance of different module configurations using both 3D and BEV AP on the 'Sedan' class, which exhibits consistent representation throughout the dataset. We begin with the Radar-only (R) configuration, where no camera-based feature refinement is applied. Subsequently, we incorporate the lifted 3D query cross-attention and adaptive fusion module (R+C), yielding an approximately 8\% improvement in both metrics. Furthermore, we integrate the azimuth-elevation-based weighting network (R+C+W), which redistributes features along the elevation domain based on their true distribution in the original Radar spectrum. This addition naturally provides greater improvement to $AP_{3D}$ with a 1.77\% gain, as it facilitates elevation refinement, while improving $AP_{BEV}$ by 0.36\%. Finally, we replaced the DINOv3 ViT with a pretrained ResNet50~\cite{he_resnet_2016} (R+C*+W), fine-tuning only the final layer. This substitution results in approximately a 3\% performance decrease in both metrics, demonstrating the superior feature representation capabilities of the DINOv3 VFM.

\begin{table}[htbp]
    \centering
    \caption{Ablation Study}
    \begin{tabular}{ccccc}
    \hline
    Metric & R & R+C & R+C+W &  R+C*+W \\ \hline \hline
    $AP_{3D}$ & 61.65 & 69.61 & \textbf{71.38} & 68.43 \\
    $AP_{BEV}$ & 66.68 & 74.96 & \textbf{75.32} & 72.42 \\
    \hline
    \end{tabular}
    \label{tab:ablation_study}
\end{table}
\section{Discussion}
\label{sec:results}
\subsection{Camera Images in Adverse Weather Conditions}
\begin{figure}[t]
  \centering
  \includegraphics[width=\linewidth]{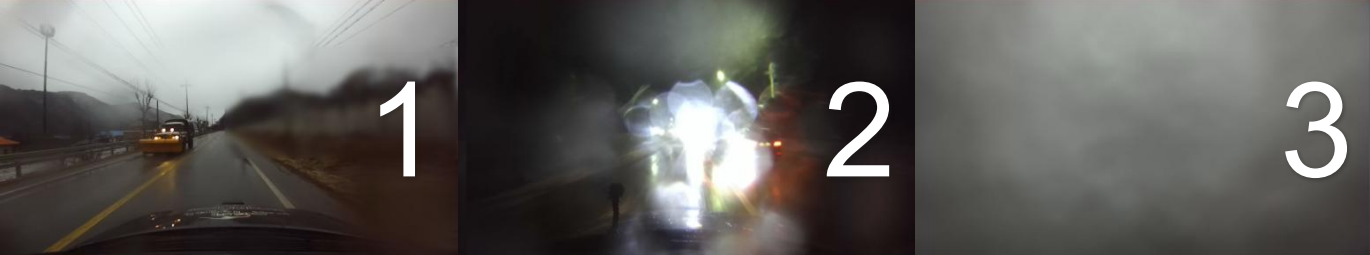}
  \caption{Examples for partially occluded (1), heavily occluded (2), and fully occluded (3).}
  \label{fig:occluded_images}
\end{figure}

Upon examination of the dataset, we discovered a substantial number of frames where the camera image is partially, heavily, or fully occluded by adverse weather effects, including rain, fog, sleet, and snow. We estimate 7.5k partially occluded, 4.8k heavily occluded, and 4.9k fully occluded frames within the whole dataset (35k frames) and present three examples in Fig.~\ref{fig:occluded_images}. These weather-related effects pose a significant challenge to the training pipeline, likely contributing to the suboptimal performance of camera-based approaches in existing methods, where the incorporation of camera data yields limited performance improvements.  

\subsection{Dataset Annotation}
The K-Radar dataset has undergone several iterations to improve annotation completeness and address visibility limitations. Version 2.0 addressed missing labels through an automated labeling pipeline utilizing LiDAR and camera data~\cite{sun_k-radar-labeling_2024}, while versions 1.1 and 2.1 provide additional information regarding the physical visibility for Radar and LiDAR modality due to sensor placement~\cite{paek_kradar_2022}. However, missing annotations in the dataset still pose a challenge for model training and evaluation. Upon examination of false positive predictions, we identified instances where our model correctly detects objects in the scene that lack corresponding ground truth bounding boxes. This phenomenon is particularly prevalent in scenarios involving parked vehicles and oncoming traffic. Consequently, this negatively affects both evaluation metrics and training dynamics, as the model receives incorrect penalty for accurate predictions, thereby degrading the learning process.

\section{Conclusion}
\label{sec:conclusion}

We present DinoRADE, a Radar-camera fusion framework for 3D object detection that performs consistently across all weather conditions and leverages inter-perspective feature aggregation using the output of a DINOv3 VFM, which improves detection especially in good weather conditions. We demonstrate the performance improvement due to cross-attention fusion with VFM image features compared to ResNet features and Radar-only detection, as well as the improvement attributed to our weighted feature lifting module for $AP_{3D}$. We present qualitative and quantitative results where we outperform existing methods by 12.1\% in $AP_{3D}$ on the 'Sedan' class. To emphasize the relevance of VRUs we report and discuss comprehensive evaluation results of multiple road users in different weather conditions where we achieve 51\% $AP_{3D}$ for 'Pedestrian' in foggy scenarios. Additionally, we compare our superior multi-class results with two baselines. With this work, we hope to further drive weather-robust object detection while providing a foundation to include performance metrics of vulnerable road user detection in future research.

\section*{Acknowledgment}
This work is supported by Infineon Technologies Austria AG and the European Union through the Horizon Europe programme (Grant Agreement project 101092834). Funded by the European Union.

\end{document}